\documentclass[runningheads]{llncs}
\usepackage[T1]{fontenc}
\usepackage{graphicx}
\usepackage{multirow}
\usepackage{subcaption}
\usepackage{amsmath,amsfonts,amssymb}
\usepackage{algorithm}%
\usepackage[table,xcdraw]{xcolor}
\usepackage{listings}
\usepackage[bookmarks=false]{hyperref}
\usepackage{orcidlink}

\usepackage[many]{tcolorbox}
 \newtcolorbox{mydoublebox}[1][]{enhanced,
  colframe=blue!75!black,
  width=\linewidth,
}

\usepackage[indLines=false]{algpseudocodex}%
\algrenewcommand\algorithmicrequire{\textbf{Input:}}

 \mathchardef\mhyphen="2D

\begin{document}
\title{Bridging Domain Knowledge and Process Discovery Using Large Language Models\thanks{\scriptsize This research was supported by the research training group ``Dataninja'' (Trustworthy AI for Seamless Problem Solving: Next Generation Intelligence Joins Robust Data Analysis) funded by the German federal state of North Rhine-Westphalia.}}

\titlerunning{Bridging Domain Knowledge and Process Discovery Using LLMs}
%
\author{Ali Norouzifar \inst{1}\orcidlink{0000-0002-1929-9992} \and Humam Kourani\inst{2, 1}\orcidlink{0000-0003-2375-2152} \and Marcus Dees\inst{3}\orcidlink{0000-0002-6555-320X} \and
Wil van der Aalst\inst{1}\orcidlink{0000-0002-0955-6940}}
\authorrunning{Ali Norouzifar et al.}
%
\institute{RWTH University, Aachen, Germany \\ \email{\{ali.norouzifar, wvdaalst\}@pads.rwth-aachen.de} \and Fraunhofer FIT, Sankt Augustin, Germany \\ \email{humam.kourani@fit.fraunhofer.de} \and
UWV Employee Insurance Agency, Amsterdam, Netherlands \\
\email{marcus.dees@uwv.nl}}
\maketitle              
\setcounter{footnote}{0}
\begin{abstract}
Discovering good process models is essential for different process analysis tasks such as conformance checking and process improvements. Automated process discovery methods often overlook valuable domain knowledge. This knowledge, including insights from domain experts and detailed process documentation, remains largely untapped during process discovery. This paper leverages Large Language Models (LLMs) to integrate such knowledge directly into process discovery. We use rules derived from LLMs to guide model construction, ensuring alignment with both domain knowledge and actual process executions. By integrating LLMs, we create a bridge between process knowledge expressed in natural language and the discovery of robust process models, advancing process discovery methodologies significantly. To showcase the usability of our framework, we conducted a case study with the UWV employee insurance agency, demonstrating its practical benefits and effectiveness.
\end{abstract}
\section{Introduction}
Recorded event data within information systems provides a rich source of information for process mining applications, enabling organizations to gain insights and improve their operational processes. In the field of process mining, various automated techniques are utilized to discover descriptive models that explain process executions. Despite the development of numerous methodologies for process discovery, the task remains inherently complex and challenging \cite{AugustoCDRMMMS19}. 
Discovering process models that do not align with domain knowledge presents significant challenges, particularly when these models are intended for conformance checking and process improvement.

In addition to the extracted event data from information systems, we often have access to domain experts, process documentation, and other resources collectively referred to as \emph{domain knowledge}, which cannot be directly used for process discovery. These valuable resources typically remain untapped when aiming to discover process models. Incorporating domain knowledge into the discovery of process models poses several challenges. For instance, domain experts usually have a thorough understanding of their processes, but they can only explain them in natural language. Furthermore, textual process documents, although potentially rich in detail, also pose integration difficulties. In our paper, we address these challenges by enabling the direct involvement of such information in process discovery through the use of Large Language Models (LLMs). LLMs have demonstrated the ability to handle user conversations and comprehend human reasoning effectively.

Our framework builds upon the IMr framework proposed in \cite{RCIS-TBP}. IMr is an inductive mining-based framework that recursively selects the process structure that best explains the actual process. Within this framework, the algorithm encounters various possibilities for constructing the process structure. To guide this selection, rules are introduced as inputs to prune the search space and eliminate potentially suboptimal process structures. Although the concept of rules is broad, the Declare rule specification language is proposed as an example~\cite{DBLP:conf/caise/MaggiBA12}. Declarative rules are advantageous due to their similarity to human reasoning and logic, supported by extensive literature. They are based on logical statements and have specific semantics, though it is unrealistic to expect users to provide these rules directly.

\begin{figure}[tb]
    \centering
    \includegraphics[width=0.65\linewidth]{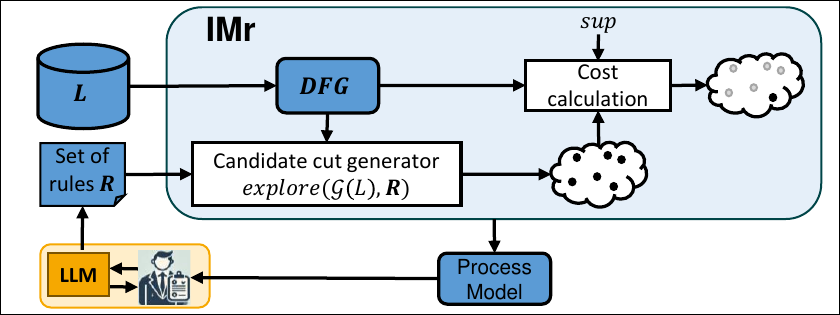}
    \caption{Our proposed framework to integrate process knowledge in the IMr framework employing LLMs.}
    \label{IMr+LLM_framework}
    \vspace{-30pt}
\end{figure}

Our proposed framework, illustrated in Fig.~\ref{IMr+LLM_framework}, leverages LLMs and prompt engineering to integrate domain knowledge into process discovery. Starting with an event log, it employs process knowledge in various forms. LLMs play a crucial role by translating textual inputs into declarative rules, which IMr then integrates. This framework allows for the use of textual process descriptions prior to initiating process discovery, enables domain experts to provide feedback on the discovered models, and facilitates interactive conversations with domain experts to gather information and improve the models.

\section{Related Work}
In traditional process discovery, event data are often used as the primary source of information to create process models~\cite{AugustoCDRMMMS19}. However, additional information resources, such as various forms of process knowledge, can significantly enhance the quality of the discovered models~\cite{DBLP:journals/cii/SchusterZA22}. When available, this supplementary knowledge can be utilized before discovery to filter the event log~\cite{DBLP:conf/caise/EckLLA15}, during the discovery phase to influence the process model structure~\cite{RCIS-TBP}, or within an interactive framework~\cite{DBLP:conf/er/DixitVBA18,DBLP:journals/softx/SchusterZA23}. Despite these benefits, the direct involvement of process experts is often limited due to the complexities involved in integrating their knowledge into process discovery.

In \cite{RCIS-TBP}, declarative rules are used as an additional input for process discovery, which can be provided by the user or generated by automated methods. However, expecting users to be proficient in declarative rule specification language is not always feasible. The proposed method in \cite{DBLP:conf/er/DixitVBA18} requires users to engage at a low level to position transitions and places based on guiding visualizations. The approach in \cite{DBLP:journals/softx/SchusterZA23} begins with an initial model discovered from a user-selected subset of variants and incrementally allows adding more variants to update the process model. Some research focuses on repairing process models after discovery, primarily to improve the correspondence between the process models and event logs, rather than incorporating process knowledge~\cite{DBLP:journals/tosem/PolyvyanyyAHW17}. In contrast, our paper aims to minimize the effort required from domain experts by using natural language conversations to influence process discovery.

Translating the natural language to process models using natural language processing is investigated in \cite{DBLP:conf/caise/AaCLR19}. Anomaly detection is examined in \cite{DBLP:journals/is/AaRL21} by focusing on semantic inconsistencies in event labels within event logs, utilizing natural language processing to identify anomalous behavior. Recently, LLMs have been employed for various process mining tasks.
The opportunities, strategies, and challenges of using LLMs for process mining and business process management are discussed in~\cite{DBLP:conf/bpm/VidgofBM23}. Additionally, several studies propose the extraction of process models directly from textual inputs~\cite{DBLP:conf/bpm/GrohsAER23,DBLP:conf/bpm/KlievtsovaBKMR23,powl_modeling}. Unlike these approaches, our method maintains the event log as the main source of information while incorporating textual process knowledge into the discovery process.

\vspace{-10pt}
\section{Background}
The blue box in Fig.~\ref{IMr+LLM_framework} highlights one recursion of the IMr framework~\cite{RCIS-TBP}. Each recursion extracts a Directly Follows Graph (DFG) from the event log, representing the set of activities $\Sigma$ and their direct succession. The algorithm searches for all binary cuts that divide $\Sigma$ into two disjoint sets considering a structure specification type, i.e., sequence, exclusive choice, concurrent, or loop type. 
IMr filters out candidate cuts that may violate any rule $r \in R$, where $R$ is the set of rules given by the user or discovered using automated methods. While \cite{RCIS-TBP} incorporates declarative constraints listed in Table.~\ref{declare-templates}, the framework is flexible to support other rule specification languages. 
Cost functions evaluate the quality of candidate cuts, based on counting the number of deviating edges and estimating the number of missing edges considering parameter $sup \in [0,1]$. In each recursion, the algorithm selects the cut with the minimum cost, splits the event log accordingly, and recursively processes each sub-log until only base cases remain.

\begin{table}[t]
    \centering
    \caption{\small Declarative templates supported by IMr~\cite{RCIS-TBP}.}
    \label{declare-templates}
    \begin{tabular}{|m{4cm}|m{8cm}|}
        \hline
        \textbf{Declarative Template} & \textbf{Description} \\
        \hline
        $at \mhyphen most(a)$ & $a$ occurs at most once. \\
        \hline
        $existence(a)$ & $a$ occurs at least once. \\
        \hline
        $response(a,b)$ & If $a$ occurs, then $b$ occurs after $a$. \\
        \hline
        $precedence(a,b)$ & $b$ occurs only if preceded by $a$. \\
        \hline
        $co\mhyphen existence(a,b)$ & $a$ and $b$ occur together. \\
        \hline
        $not\mhyphen co\mhyphen existence(a,b)$ & $a$ and $b$ never occur together. \\
        \hline
        $not\mhyphen succession(a,b)$ & $b$ cannot occur after $a$. \\
        \hline
        $responded\mhyphen existence(a,b)$ & If $a$ occurs in the trace, then $b$ occurs as well. \\
        \hline
    \end{tabular}
    \vspace{-10pt}
\end{table}

\vspace{-10pt}
\section{Motivating Example}
To motivate the research question addressed in this paper, consider the following event log extracted from a synthetic process {\small$L{=}[\langle$ A-created, A-canceled $\rangle^{300}$, 
      $\langle$ A-created, Doc-checked, Hist-checked, A-accepted $\rangle^{200}$,
    $\langle$ A-created, Hist-checked, Doc-checked, A-accepted $\rangle^{50}$, 
      $\langle$ A-created, Doc-checked, Hist-checked, A-rejected $\rangle^{300}$,
      $\langle$ A-created, Hist-checked, Doc-checked, A-rejected $\rangle^{80}$, 
      $\langle$ A-created, A-canceled, A-accepted$\rangle^{20}$, 
      $\langle$ A-created, A-canceled, A-rejected $\rangle^{15}$, 
      $\langle$ A-created, Doc-checked, Hist-checked, A-rejected, A-accepted~$\rangle^{35}]$}, where A stands for application, Doc for documents, and Hist for history.

\begin{figure}[htb]
\centering 
\begin{subfigure}{1\textwidth}
\centerline{\includegraphics[scale=0.10]{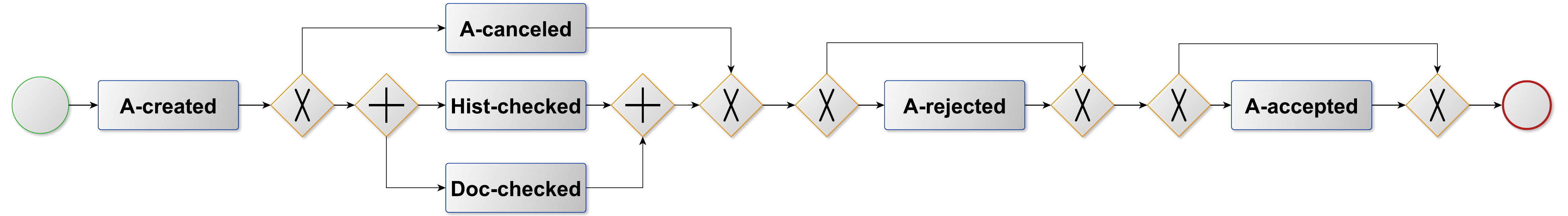}}
 \caption{\footnotesize Discovered model with deviations from the process knowledge. This model is discovered using IMf with $f{=}0.2$ and IMr with $sup{=}0.2$ and rules discovered employing Declare Miner~\cite{DBLP:conf/caise/MaggiBA12} with $confidence{=}1$.}
 \label{motiv-imf-imr}
 \end{subfigure}\\
 \begin{subfigure}{1\textwidth}
 \centerline{\includegraphics[scale=0.10]{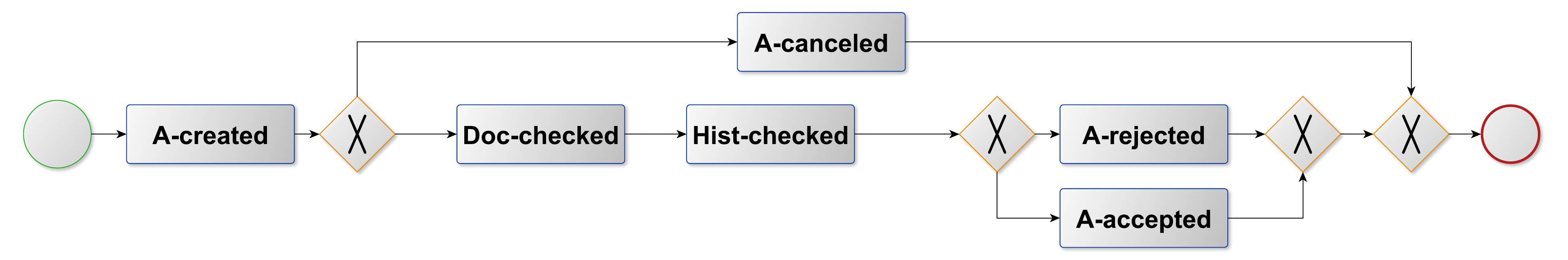}}
\caption{\footnotesize The desired process model considering both the event log and process knowledge. IMr with $sup{=}0.2$ and rules extracted from process description employing ChatGPT discover this model.}
\label{motiv-imr+LLM}
\end{subfigure}
\caption{\small Discovered models from the motivating example event log using different techniques.}
\end{figure}

Figure \ref{motiv-imf-imr} illustrates the process model discovered using the IMf algorithm as a state-of-the-art process discovery technique~\cite{leemans2013discoveringIMF}. The IMr framework with parameter $sup=0.2$ and utilizing the Declare Miner~\cite{DBLP:conf/caise/MaggiBA12} with $confidence=1$ discovers the same process model. Consider that in addition to the provided event log, we have some additional process knowledge that helps us verify this model and pinpoint the possible unexpected behavior represented in the process model. In this paper, ChatGPT refers to ChatGPT-4o. We provided a text as feedback on this discovered model and asked ChatGPT to translate natural language feedback into understandable rules for the IMr framework. Here is our written feedback:

 \begin{tcolorbox}[enhanced,breakable,colback=gray!5!white,colframe=blue!75!black,left=0.5mm, right=0.5mm]
{\small
The discovered process does not fully adhere to our intuitions. Specifically, if a claim is canceled, the application cannot be either rejected or accepted. Furthermore, a claim cannot be both rejected and accepted for a single individual. Additionally, the history is always checked after the documents have been reviewed.
}
 \end{tcolorbox}

The following declarative rules, as explained in this paper, were extracted by ChatGPT:
  \begin{tcolorbox}[enhanced,breakable,colback=gray!5!white,colframe=red!75!black,left=0.5mm, right=0.5mm]
{\small
not-co-existence(A-cancelled,A-accepted),
not-co-existence(A-cancelled,A-rejected),
not-co-existence(A-accepted,A-rejected),
response(Doc-checked,Hist-checked)
}
 \end{tcolorbox}

Figure \ref{motiv-imr+LLM} presents the process model discovered using our proposed pipeline. In this approach, we utilized ChatGPT to interpret the textual feedback and generate declarative constraints, which are then used as input for the IMr framework. 
\vspace{-20pt}

\vspace{-40pt}
\section{Domain-Enhanced Process Discovery with LLMs}\label{sec:framework}
\begin{figure}[htb]
    \centering
    \vspace{-20pt}
\includegraphics[width=0.5\linewidth]{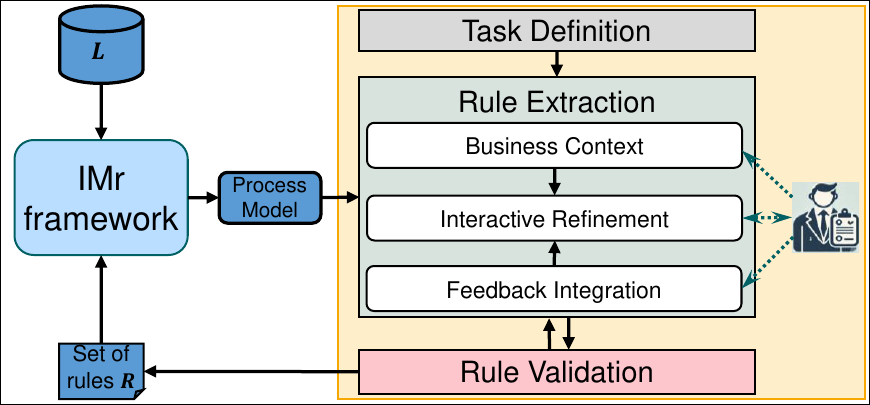}
    \caption{\small Different components of the designed framework to bridge domain knowledge and process discovery using LLMs.}
    \label{fig:prompts}
    \vspace{-20pt}
\end{figure}

In this section, we present our framework that leverages LLMs to integrate domain knowledge into the process discovery task. \autoref{fig:prompts} illustrates an overview of our proposed framework. The core idea is to utilize domain knowledge to generate a set of rules $R$ which serves as input for the IMr framework. This can be done before starting the discovery by encoding process descriptions as rules, or after the process discovery by having a domain expert review the process model and provide feedback. Engaging in interactive conversations with LLMs in both scenarios helps address uncertainties and improve the quality of the extracted rules. 
An implementation of the framework is publicly available\footnote{\url{https://github.com/aliNorouzifar/IMr-LLM.git}}.

\vspace{-10pt}
\subsection{Task Definition}
As outlined in \cite{powl_modeling}, role promoting, knowledge injection, few-shot learning, and negative prompting techniques have significant potential to effectively prepare LLMs for specific process mining tasks. In our initial prompt, we define the role of the LLM as an interface between the domain expert and process discovery framework, such that LLM should encode the domain knowledge to declarative constraints as we need in IMr. Despite the similarity of declarative templates to human logic and reasoning, we observed the difficulties of LLMs in adhering to strict expectations. Therefore, we explain in our prompt the set of constraints we support, detailing both the syntax and the semantics of these constraints (cf. \autoref{declare-templates}). We leverage the LLM's ability to derive insights from examples by providing multiple pairs of textual process descriptions and their corresponding declarative constraints. Additionally, we include instructions to avoid common issues, such as syntactic mistakes, and extend our learning pairs to include examples of undesirable constraints. 
The detailed written prompt is available in our GitHub repository\footnote{\url{https://github.com/aliNorouzifar/IMr-LLM/blob/master/files/prompts.pdf}}.

\vspace{-10pt}
\subsection{Rule Extraction}
After introducing the task, the LLM is ready to receive textual input and produce output as declarative constraints. As illustrated in Fig.~\ref{fig:prompts}, domain experts can contribute in three distinct ways: providing business context, offering feedback after reviewing process models, and engaging in interactive conversations with the LLM. In the following sections, we explain these contributions in detail and discuss their respective roles.

\vspace{-10pt}
\subsubsection{Business Context}
The domain expert can introduce the actual business process to the LLM, providing a general overview, detailing the relationships between specific activities, or even including constraints written in natural language. This flexibility allows the domain expert to tailor the input based on their unique insights and the specifics of the process at hand. It is important to note that the LLM is unaware of specific activity labels used in the recorded event data. The list of activities can be automatically derived from the event log, ensuring that all relevant actions are accurately captured in the generated constraints. Alternatively, the domain expert can provide the list of activities and add context to guide the LLM in relating the process description with the activity labels, resulting in constraints that involve the correct activity labels.

\vspace{-10pt}
\subsubsection{Feedback integration}
After generating the initial process model, it is presented to the domain expert for review. 
The domain expert is expected to examine the process model for accuracy, completeness, and practical alignment with real-life scenarios. In case of finding errors in the represented model, the domain expert can provide a written feedback and explain the behaviors that do not make sense in the real process.
The LLM then adjusts and refines the declarative constraints based on this feedback. 

\subsubsection{Interactive Refinement} 
In typical scenarios, LLMs tend to provide answers that appear confident and definitive, often without indicating any uncertainty~\cite{DBLP:journals/corr/abs-2311-05232}. We facilitate a more detailed understanding of the provided textual descriptions by encouraging the LLM to express uncertainty and address it by asking questions. 
This stage involves a dynamic dialogue between the LLM and the domain expert. 
Should it encounter gaps in its knowledge or find ambiguities in the process descriptions, the LLM is encouraged to formulate and pose relevant questions. These questions are directed towards the domain experts, who then provide responses. The quality and precision of the responses provided by domain experts play a significant role in enhancing the quality of the generated constraints.

\vspace{-10pt}
\subsection{Rule Validation}  An essential step in the framework is checking the extracted declarative constraints from the LLM's response. The LLM is instructed to encapsulate the constraints within specific tags in the response and to write them in a predefined language with no additional text or descriptions. Following extraction, the constraints undergo a validation process. This includes checking that the syntax of each constraint conforms to our predefined language, e.g., checking the type identifier and the number of activities specified within the constraint. Additionally, the labels of activities are verified against the activities recorded in the event log. If any errors are detected during validation, an error-handling loop is initiated. A new prompt specifies the problem and its location, prompting the LLM to adjust its output. 

\vspace{-10pt}
\section{Case Study}
\begin{figure}
    \centering
    \vspace{-20pt}
\includegraphics[width=1\linewidth]{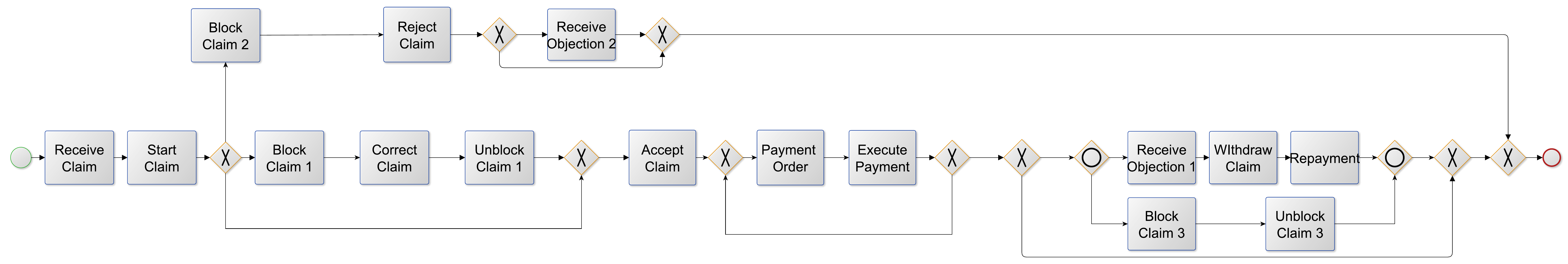}
    \caption{\small Normative model of the UWV claim handling process, extracted manually in collaboration with domain experts \cite{RCIS-TBP}.}
    \label{normativ_UWV}
    \vspace{-20pt}
\end{figure}

A case study with the UWV employee insurance agency is conducted to demonstrate the usability of our approach in a real-life setting. UWV is responsible for managing unemployment and disability benefits in the Netherlands. For this case study, one of UWV's claim-handling processes is selected. Figure~\ref{normativ_UWV} depicts the normative model of this process, which was developed in collaboration with process experts who have a thorough understanding of the workflow. The event log used in this study contains 144,046 cases, 16 unique activities, and 1,309,719 events. Our GitHub repository provides the full prompting history and more readable process models\footnote{\url{https://github.com/aliNorouzifar/IMr-LLM/blob/master/files/prompts.pdf}}.

\vspace{-10pt}
\subsection{Process Discovery Without Including Process Knowledge}
\begin{figure}[htb]
\centering 
 \begin{subfigure}{1\textwidth}
 \centerline{\includegraphics[scale=0.09]{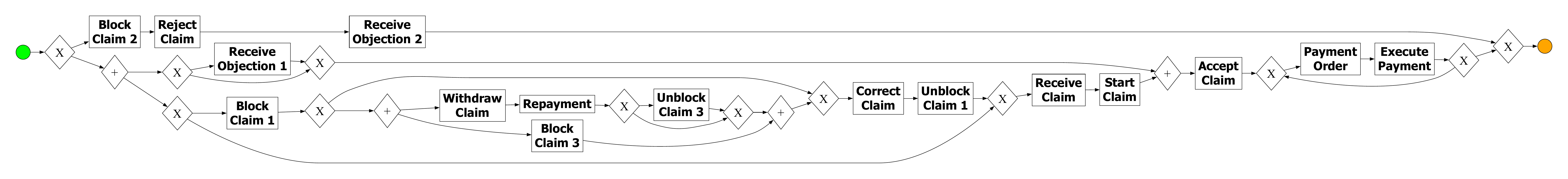}}
\caption{\footnotesize Discovered model with using IMf with $f=0.2$.}
\label{UWV_imf}
\end{subfigure}\\
\begin{subfigure}{1\textwidth}
\centerline{\includegraphics[scale=0.09]{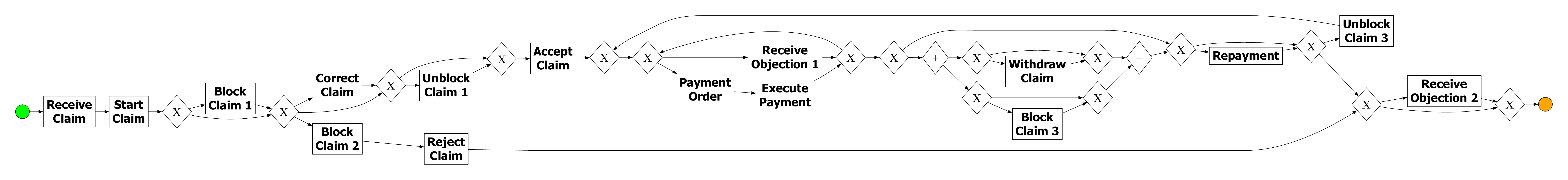}}
\caption{\footnotesize Discovered model using IMr with $sup=0.2$ without any rules.}
\label{UWV_imr+LLM0}
\end{subfigure}\\
 \begin{subfigure}{1\textwidth}
\centerline{\includegraphics[scale=0.1]{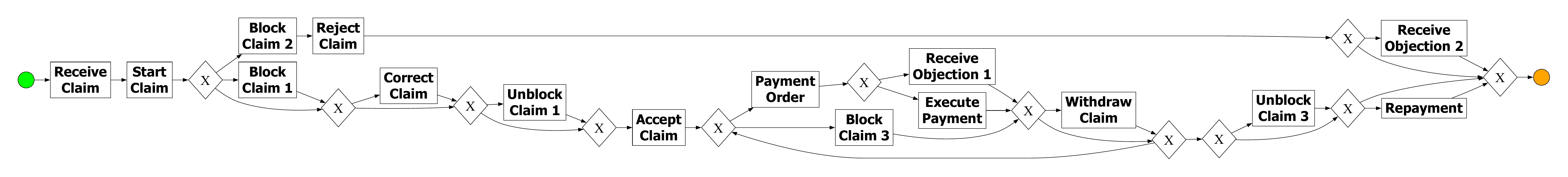}}
\caption{\footnotesize Discovered model using IMr with $sup=0.2$ and the rules provided by ChatGPT.}
\label{UWV_imr+LLM1}
\end{subfigure}\\
 \begin{subfigure}{1\textwidth}
\centerline{\includegraphics[scale=0.09]{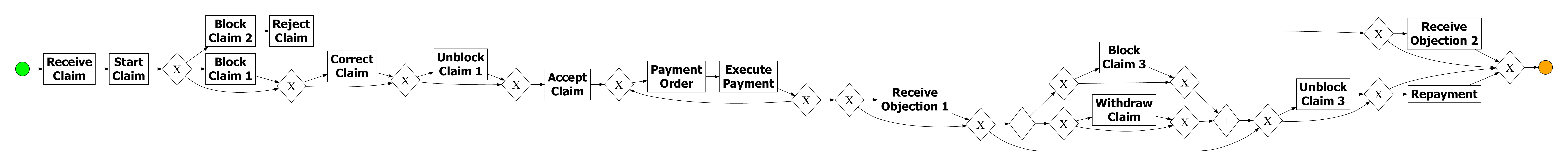}}
\caption{\footnotesize Discovered model using IMr with $sup=0.2$ and the rules provided by ChatGPT after incorporating the domain expert feedback.}
\label{UWV_imr+LLM2}
\end{subfigure}
\caption{\small Discovered models from UWV event log using different strategies.}
\vspace{-20pt}
\end{figure}

Our initial attempt to discover a process model using the IMf algorithm with $f=0.2$ resulted in the model shown in Fig.~\ref{UWV_imf}. When compared to the normative model, significant differences are observed, e.g., \textit{Receive Claim} and \textit{Start Claim} are the first mandatory steps but the process model allows for skipping them or for many other activities occurring before them. Fig.~\ref{UWV_imr+LLM0} illustrates the process model discovered using the IMr algorithm with $sup=0.2$ and an empty set of input rules. Although this model shows more structural similarities to the normative model, it still contains some nonsensical differences. For instance, \textit{Block Claim 1} should only be relevant if the claim is planned to be accepted, but this model permits it for rejected cases as well. Similarly, \textit{Receive Objection~2} should only occur if the claim is rejected, yet the model allows it for accepted cases as well. 
\vspace{-20pt}

\subsection{Employing ChatGPT to Extract the Rules}
We experimented with Gemini and various versions of ChatGPT to translate the process knowledge into declarative rules. ChatGPT-4o provided the best constraints and demonstrated a superior understanding of the task. By incorporating rules extracted by ChatGPT into the IMr framework, we obtained the process model shown in Fig.~\ref{UWV_imr+LLM1}. After iterating with feedback from domain experts, the model is further refined and improved, resulting in the process model depicted in Fig.~\ref{UWV_imr+LLM2}.

 The process knowledge for this case study is incorporated as a set of activity labels and a process description provided by a domain expert. The blue boxes indicate the prompts given by domain experts, while the red boxes display the responses generated by ChatGPT.

\begin{tcolorbox}[enhanced,breakable,colback=gray!5!white,colframe=blue!75!black,left=0.5mm, right=0.5mm]
\small

Our process is designed to handle claim applications within an insurance agency, where applicants request various types of benefits, and our system processes these applications through a structured workflow. There are three types of blocking mechanisms: Block Claim 1, Block Claim 2, and Block Claim 3, each with corresponding resolutions: Unblock Claim 1, Unblock Claim 2, and Unblock Claim 3. The process begins once a claim is received. Block Claim 1 occurs when initial issues are detected, leading to necessary corrections before the claim is unblocked and accepted, resulting in the client receiving between one to three payments. Block Claim 2 is applied when the claim is deemed ineligible for receiving any payments, resulting in rejection and potentially an objection from the client. Block Claim 3 is triggered if there is missing information, such as incorrect bank account details, after a claim is accepted; the client is notified to provide the correct details, and the claim is unblocked upon receipt of this information. Additionally, clients may file an objection after receiving payments, leading to the withdrawal of the claim and repayment of the received money to the agency.
\end{tcolorbox}
 
  Following this process description, we encouraged ChatGPT to ask clarifying questions whenever it encountered uncertainties about the declarative rules. The box below provides an example of questions posed by ChatGPT.
 
\begin{tcolorbox}[enhanced,breakable,colback=gray!5!white,colframe=red!75!black,left=0.5mm, right=0.5mm]
\small
\textbf{Q1}: Can "Block Claim 1," "Block Claim 2," and "Block Claim 3" occur in the same process instance, or are they mutually exclusive?
\textbf{Q2}: Are the corresponding unblocking activities ("Unblock Claim 1," "Unblock Claim 2," and "Unblock Claim 3") strictly linked to their respective blocking activities, or can there be other sequences?
\textbf{Q3}: Is there any restriction on the order of "Receive Objection 1" and "Receive Objection 2," or can they occur independently of each other?
...
\end{tcolorbox}

The provided questions demonstrate the LLM's correct approach to understanding the declarative constraints between the activities in the process. For instance, the first question indicates that ChatGPT is attempting to discern whether there is a co-existence or non-co-existence relationship between block claim types. Question 3 is also pertinent, as the process description does not clearly distinguish between Receive Objection 1 and Receive Objection 2. The domain expert provided the following answers to these questions:

\begin{tcolorbox}[enhanced,breakable,colback=gray!5!white,colframe=blue!75!black,left=0.5mm, right=0.5mm]
\small
\textbf{A1}:  Some cases may have both Block Claim 1 and Block Claim 3. But Block Claim 2 can never occur in the trace with Block Claim 1 or Block Claim 3. 
\textbf{A2}: Each blocking mechanism (Block Claim 1, Block Claim 2, Block Claim 3) has a corresponding unblocking mechanism (Unblock Claim 1, Unblock Claim 2, Unblock Claim 3), and each pair (Block, Unblock) occurs in a strict sequence.
\textbf{A3}: Receive Objection 2 only occurs in the rejected cases and Receive Objection 1 only occurs in the accepted cases after they receive some payments. In general, Receive Objection 1 and Receive Objection 2 cannot occur together.
...
\end{tcolorbox}

The set of declarative constraints extracted after answering the questions is:

\begin{tcolorbox}[enhanced,breakable,colback=gray!5!white,colframe=red!75!black,left=0.5mm, right=0.5mm]
\small
not-co-existence(Block Claim 2, Block Claim 1), 
not-co-existence(Block Claim 2, Block Claim 3), 
co-existence(Block Claim 1, Unblock Claim 1), 
co-existence(Block Claim 2, Unblock Claim 2), 
co-existence(Block Claim 3, Unblock Claim 3), 
precedence(Block Claim 1, Unblock Claim 1), 
precedence(Block Claim 2, Unblock Claim 2), 
precedence(Block Claim 3, Unblock Claim 3), 
not-co-existence(Receive Objection 1, Receive Objection 2), 
precedence(Reject Claim, Receive Objection 2), 
precedence(Payment Order, Receive Objection 1), 
at-most(Correct Claim), 
precedence(Block Claim 1, Correct Claim), 
precedence(Correct Claim, Unblock Claim 1), 
response(Withdraw Claim, Repayment), 
responded-existence(Accept Claim, Payment Order), 
responded-existence(Payment Order, Execute Payment)
\end{tcolorbox}

These rules are validated using a predefined program to ensure that the activity labels are consistent with the event log and that the declarative constraints are free of syntax errors. Then, the rules are used as input for the IMr framework in addition to the event log, and the process model represented in Fig.~\ref{UWV_imr+LLM1} is discovered. These rules are aligned with the process description and the follow-up questions and answers. 
For example, the answer to the first question (A1) led to the extraction of \textit{not-co-existence(Block Claim 2, Block Claim 1)} and \textit{not-co-existence(Block Claim 2, Block Claim 3)}, correctly illustrating the relationship between these activities. These rules help IMr avoid the incorrect positioning of \textit{Block Claim 1} observed in Fig.~\ref{UWV_imr+LLM0}. Another improvement is achieved by considering \textit{not-co-existence(Receive Objection 1, Receive Objection 2)}, which prevents \textit{Receive Objection 1} and \textit{Receive Objection 2} from occurring in the same trace. Fig.~\ref{UWV_imr+LLM0} allows for \textit{Block Claim 3} without the existence of \textit{Unblock Claim 3}. The rule \textit{co-existence(Block Claim 3, Unblock Claim 3)} guides IMr to avoid placing \textit{Unblock Claim 3} as the re-do part of a loop. We presented this process model to domain experts for feedback. They identified some potential issues, which are then provided to ChatGPT to generate a better set of declarative templates.

\begin{tcolorbox}[enhanced,breakable,colback=gray!5!white,colframe=blue!75!black,left=0.5mm, right=0.5mm]
\small
The discovered process model is interesting but we observe some issues. For example, Receive Objection 1 should occur after all the payments are executed. This time of objection can occur at most one time per claim. Withdraw Claim can not be followed by another payment. Usually, after the payments are executed, the applicant has the option to send an objection, withdraw the claim, and repay the received benefits. Withdraw Claim only occurs at most once per claim.
 \end{tcolorbox}

After the consideration of the domain expert input, these constraints are added by ChatGPT to the previous set of constraints:

\begin{tcolorbox}[enhanced,breakable,colback=gray!5!white,colframe=red!75!black,left=0.5mm, right=0.5mm]
\small
precedence(Execute Payment, Receive Objection 1), 
at-most(Receive Objection 1), 
not-succession(Withdraw Claim, Payment Order), 
not-succession(Withdraw Claim, Execute Payment), 
at-most(Withdraw Claim)
\end{tcolorbox}

The discovered model using the modified set of constraints is illustrated in Fig.~\ref{UWV_imr+LLM2}. While this model differs from the normative model in Fig.~\ref{normativ_UWV}, it better represents the actual process compared to the models in Fig.~\ref{UWV_imf} and Fig.~\ref{UWV_imr+LLM0}, which were discovered without considering process knowledge. In comparison to Fig.~\ref{UWV_imr+LLM1} some improvements are achieved considering the provided feedback. The constraint \textit{at-most(Receive Objection 1)} prevents \textit{Receive Objection 1} from being included in a loop, and \textit{precedence(Execute Payment, Receive Objection 1)} ensures it is positioned after \textit{Execute Payment}. Additionally, \textit{Withdraw Claim} is no longer in a loop due to the \textit{at-most(Withdraw Claim)} constraint and is correctly positioned after \textit{Payment Order} and \textit{Withdraw Claim} because of the constraints \textit{not-succession(Withdraw Claim, Payment Order)} and \textit{not-succession(Withdraw Claim, Execute Payment)}.

\vspace{-10pt}
\section{Conclusion}
The integration of process knowledge in the discovery of process models is often overlooked in the literature. In this paper, we leveraged advancements in LLMs to demonstrate their capabilities in encoding textual domain knowledge into comprehensible rules for process discovery. Our proposed framework not only facilitates the integration of feedback from domain experts but also enables interactive improvement of process models. Through a comprehensive case study, we demonstrated the effectiveness of our framework in generating process models that better align with process knowledge. While the extracted set of declarative constraints from LLMs shows great promise, there is still room for improvement in precision and completeness. Future work focuses on expanding the range of declarative templates within the IMr framework and developing additional rule specification patterns. Additionally, providing more detailed examples in task definition steps helps LLMs capture a broader context, further enhancing the quality of the extracted constraints.

\vspace{-10pt}
\bibliographystyle{splncs04}
\bibliography{lit}

\end{document}